%% file: template.tex
\documentclass{article}

\usepackage{arxiv}

\usepackage[utf8]{inputenc} 
\usepackage[T1]{fontenc}    
\usepackage[hyphens]{url}
\usepackage[colorlinks = true,
            linkcolor = blue,
            urlcolor  = blue,
            citecolor = blue,
            anchorcolor = blue]{hyperref}

\usepackage{booktabs}       
\usepackage{amsfonts}       
\usepackage{nicefrac}       
\usepackage{microtype}      
\usepackage{lipsum}

\usepackage{float}
\usepackage{color}
\usepackage{tabularray}
\usepackage{array}
\usepackage{multirow}
\usepackage{ragged2e}
\usepackage{booktabs}
\usepackage{vcell}
\usepackage{subcaption}
\usepackage{amsmath,amssymb,amsfonts}
\usepackage{algorithmic}
\usepackage{graphicx}
\usepackage{textcomp}
\usepackage{booktabs}
\usepackage{multirow}
\usepackage{subcaption}
\usepackage{caption}
\usepackage{siunitx}
\usepackage{pifont}
\usepackage{tipa}
\usepackage{tfrupee}
\usepackage{soul}
\usepackage{adjustbox}
\usepackage{tabularx,ragged2e,booktabs}

\usepackage{graphicx}

\title{A Comparative Study of Machine Learning Algorithms for Anomaly Detection in Industrial Environments: Performance and Environmental Impact}

\author{%
  \href{https://orcid.org/0000-0003-2165-0144}{\includegraphics[scale=0.06]{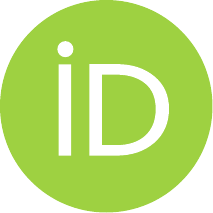}\hspace{1mm}{\'A}lvaro Huertas-Garc{\'i}a},
  \href{https://orcid.org/0009-0003-1387-1630}{\includegraphics[scale=0.06]{orcid.pdf}\hspace{1mm}Carlos Mart{\'i}-Gonz{\'a}lez},
  \href{https://orcid.org/0000-0001-9500-3121}{\includegraphics[scale=0.06]{orcid.pdf}\hspace{1mm}Rub{\'e}n Garc{\'i}a Maezo},
  \href{}{\includegraphics[scale=0.06]{orcid.pdf}\hspace{1mm}Alejandro Echeverr{\'i}a Rey} \\
  \vspace{0.5em}
  Department of Artificial Intelligence \\
  FUNDITEC \\
  C/ Faraday 7, Campus de Cantoblanco \\
  28049 Madrid \\
  \texttt{\{ahuertas, cmarti, rugarcia, aecheverria\}@funditec.es}
}

\renewcommand{\shorttitle}{\textit{arXiv} Template}

\hypersetup{
pdftitle={DETECTA-PHASE 1},
pdfsubject={q-bio.NC, q-bio.QM},
pdfauthor={David S.~Hippocampus, Elias D.~Striatum},
pdfkeywords={First keyword, Second keyword, More},
}

\begin{document}
\maketitle

\begin{abstract}

In the context of Industry 4.0, the use of artificial intelligence (AI) and machine learning for anomaly detection is being hampered by high computational requirements and associated environmental effects. This study seeks to address the demands of high-performance machine learning models with environmental sustainability, contributing to the emerging discourse on 'Green AI.' An extensive variety of machine learning algorithms, coupled with various Multilayer Perceptron (MLP) configurations, were meticulously evaluated. Our investigation encapsulated a comprehensive suite of evaluation metrics, comprising Accuracy, Area Under the Curve (AUC), Recall, Precision, F1 Score, Kappa Statistic, Matthews Correlation Coefficient (MCC), and F1 Macro. Simultaneously, the environmental footprint of these models was gauged through considerations of time duration, CO2 equivalent, and energy consumption during the training, cross-validation, and inference phases. Traditional machine learning algorithms, such as Decision Trees and Random Forests, demonstrate robust efficiency and performance. However, superior outcomes were obtained with optimised MLP configurations, albeit with a commensurate increase in resource consumption. The study incorporated a multi-objective optimisation approach, invoking Pareto optimality principles, to highlight the trade-offs between a model's performance and its environmental impact. The insights derived underscore the imperative of striking a balance between model performance, complexity, and environmental implications, thus offering valuable directions for future work in the development of environmentally conscious machine learning models for industrial applications.
\end{abstract}

\keywords{Anomaly Detection\and Green AI \and 
Trustworthy AI \and Machine Learning \and Artificial Intelligence \and Industrial Environments \and Comparative Study \and Environmental Impact}

\section{Introduction}

The ongoing digital transformation has been revolutionizing various sectors of the economy, including the industrial sector. This transformation, often referred to as Industry 4.0, has engendered increasingly dynamic, interconnected, and inherently complex manufacturing environments~\cite{review_AI_industry_2020}. The key to this transformation is the generation of large volumes of data collected through numerous sensors embedded in industrial processes. Data collected from these sources, when used appropriately, can lead to remarkable improvements in process monitoring, optimization, equipment integrity, and worker safety, while simultaneously reducing operational costs~\cite{review_industry_AI_2018}.

However, the sheer volume and complexity of the data produced lead to significant challenges in identifying unusual patterns or anomalies, which could potentially signal substantial problems or inefficiencies~\cite{kim_industrial_2022}. Machine learning, a subset of artificial intelligence (AI), has demonstrated its capability to effectively detect such anomalies, thus offering the potential for automated and intelligent anomaly detection systems~\cite{Audibert_2022}. Yet, the broad implementation of AI and machine learning in real manufacturing environments still faces significant hurdles, limiting its transition beyond experimental pilot stages~\cite{review_AI_industry_2020, Audibert_2022}.

One of these challenges, and the focus of this study, relates to the environmental impact of AI operations. The computational requirements of machine learning can be substantial, leading to significant energy consumption and consequent environmental implications~\cite{schwartz2019green}. This has given rise to the field of `Green AI', which emphasizes developing AI solutions that are not only effective but also environmentally friendly.

In light of this, it is crucial to consider the environmental impact of the AI systems developed to manage and optimize industrial operations alongside the environmental footprint of the industrial operations themselves. As industries worldwide strive to reduce their carbon footprint, it becomes increasingly important to develop AI-driven strategies that are both efficient and environmentally conscious.

To achieve this primary goal, specific objectives are addressed:

\begin{itemize}
    \item Processing and extraction of useful data for the formulation of anomaly detection criteria, validated in collaboration with subject-matter experts.

    \item Use of these criteria for the generation of a labeled dataset that allows the development of supervised machine learning models.

    \item Development and evaluation of machine learning models for anomaly detection and classification.
    
    \item To explore the trade-offs and synergies between algorithmic performance and environmental impact, contributing to the broader discourse on Green AI and its role in sustainable industrial practices.
    
    \item To provide insights into how machine learning can be leveraged responsibly and energy-efficiently in the industrial sector.
\end{itemize}

By addressing these objectives, this study aims to bridge the gap between the field of AI and the manufacturing industry, facilitating the transition towards Industry 4.0 supported by AI~\cite{review_AI_industry_2020}. The findings of this study are expected to assist researchers and manufacturers in understanding the requirements and steps necessary for this transition, as well as the challenges that may arise during this process.

\section{Related Works}

The recent increase in the utilization and development of Artificial Intelligence (AI) has come with an increased understanding of the environmental footprint these technologies leave behind~\cite{green_ai_smart_cities_2021, green_ai_climate_change_ml_2022}. Historically, AI research and development has been predominately focused on enhancing accuracy and performance, often overlooking energy efficiency~\cite{AI_sustainability}. However, the scenario is now changing, with the realization that energy efficiency is vital not only for environmental sustainability but also for AI technologies scalability and practical implementation~\cite{kim2021ibert}.  In fact, the computations required for AI research have been doubling every few months, resulting in a staggering 300,000x increase from 2012 to 2018~\cite{schwartz2019green}. This paradigm shift has given rise to the concept of Green AI, an initiative that advocates for the development of AI technologies that are environmentally friendly and sustainable~\cite{review_AI_industry_2020, review_industry_AI_2018}. In this context, several strategies have been proposed to make AI research and development more energy-efficient.

A pragmatic solution to this issue has been proposed by Schwartz et al.~\cite{schwartz2019green}—introducing efficiency as an evaluation criterion for research, alongside accuracy and other similar measures. By including the financial cost or "price tag" of developing, training, and running models, researchers can establish baselines for investigating increasingly efficient methods. 

Patterson et. al.~\cite{green_ai_best_practices_2022} projected that the total carbon emissions from AI model training could decline by 2030 if the entire field embraced best practices, proposing a set of practices that could potentially reduce CO2 emissions by up to a thousand times. Firstly, utilizing the most efficient processors hosted in the most environmentally-friendly datacenters, often available in the cloud. Secondly, developing more efficient models by harnessing sparsity or integrating retrieval into a smaller model. Thirdly, encouraging transparency by disclosing energy consumption and carbon footprint to stimulate competition based on parameters beyond just model quality. Lastly, employing renewable energy sources to power AI training whenever feasible.

Another noteworthy approach centers on conducting modifications on datasets. A recent exploratory study~\cite{data_centric_AI_2022} discovered that by altering datasets, energy consumption could be substantially reduced, in some cases by up to 92.16\%, often with little or no decline in model accuracy. This strategy suggests that careful and thoughtful data preprocessing and management can play a crucial role in promoting energy efficiency in machine learning.

The profiling of energy consumption for inference tasks is another viable strategy for enhancing energy efficiency. An empirical model has been developed to estimate the energy consumption of specific inference tasks on edge computing devices~\cite{energy_deep_learning_2022}. This model can guide the search for efficient neural network architectures, serve as a heuristic in neural network pruning, or assist in evaluating and comparing the energy performance of various deep neural network architectures.
Consider the case of Large Language Models (LLMs), such as GPT-3~\cite{gpt3_article_2020}, which have enabled breakthroughs in Natural Language Processing (NLP), however, they also present a significant computational and energy challenge. As a result of this scenario, model quantization has also been explored, which reduces both the memory footprint and the computational resources required and can contribute to lower energy consumption during model inference~\cite{kim2021ibert,wu2020integer}.

Lastly, comparative analyses of machine learning models also hold promise for improving energy efficiency. For instance, a study~\cite{Kanewala2021ExploringDL} that compared the use of feedforward neural networks, random forests, and recurrent neural networks for predicting the energy data of a chiller system found that best practices can enhance model performances.

Despite the recent surge in Green AI, there remain substantial gaps in the literature. The current discourse predominantly focuses on Deep Learning models and emerging trends like model quantization, often neglecting the potential of simpler, less resource-intensive tools, like traditional Machine Learning algorithms~\cite{green_ai_best_practices_2022}. This oversight is particularly evident in the context of anomaly detection in industrial environments, an area that stands to benefit significantly from energy-efficient algorithms. Furthermore, the environmental impact of various Machine Learning and Deep Learning algorithms in such settings remains underexplored.

This study aims to address these gaps by presenting a comparative analysis of different Machine Learning, Deep Learning, and quantized versions of these algorithms. We evaluate these algorithms not only on performance metrics but also on their environmental footprint. This approach extends the principles of Green AI into a practical industrial scenario, specifically in the context of anomaly detection in an environmental industry. By doing so, we aim to highlight more sustainable and efficient methodologies that could revolutionize production practices~\cite{7005202}.

In conclusion, the importance of energy efficiency in machine learning is now being recognized, and several strategies are being explored to reduce its energy consumption and environmental impact. By adopting these strategies, the machine learning field can align more closely with the principles of Green AI, promoting both environmental sustainability and practical scalability.

\section{Methodology}

\subsection{Data}
This research aims to detect and predict anomalies within an industrial milling machine using a dataset that has been carefully curated and labelled under the supervision of experts. The dataset contains instances representing 30-second windows of measurement from sensors monitoring temperature, vibrations, and current. In total, 308,772 instances were collected and presented as a three-class imbalanced problem resulting from meticulous criteria establishment and label propagation. The instances are classified according to whether they represent a non-anomalous state (99.86\%), a single sensor anomaly (0.01\%), or a multiple sensor anomaly (0.13\%).

As part of the data engineering phase, measurements were captured every second and grouped into 30-second intervals. These grouped data points were then transformed into a new set of features using descriptive statistical methods. These methods include calculations of mean, maximum, minimum, kurtosis, skewness, and the number of current peaks. This process effectively expanded the initial 7-feature dataset into one comprising 38 features, providing a more nuanced view of the machinery's functioning over each time interval. Any missing values in numerical data were managed by mean imputation, ensuring valuable information preservation. 

Due to multicollinearity, we implemented a threshold of 0.9 to prevent potential negative impacts on our machine learning models. Features with correlations above this threshold were removed, which primarily impacted current and vibration sensor readings.  Our feature set was reduced from 38 to 25, resulting in a more lean and efficient dataset for modeling. 

During the development phase of the machine learning model, we focused on the performance and reproducibility of the model, maintaining a random seed value of 1794 throughout all experiments. In order to prevent overfitting and underfitting, we used the stratified Kfold method with ten folds for cross-validation. Our data was divided so that 70\% was utilized for training and validation, with the remaining 30\% reserved as a holdout test set for final model evaluation. 

\subsection{Machine Learning Algorithms}
 
In our pursuit to decipher anomaly detection in the context of Green AI, we have sought to include a broad spectrum of machine learning algorithms, each bringing a unique perspective and capability to our investigation. The selected models span across various categories, namely linear models, non-linear models, single tree-based methods, ensemble decision trees, boosting decision trees, and deep learning models. This range allows us to glean a more holistic understanding of the data and identify different types of anomalies efficiently~\cite{Nassif2021MachineLF}.

Linear models we've implemented are Logistic Regression, Ridge Classifier, Naive Bayes, Linear Discriminant Analysis (LDA), and a Support Vector Machine (SVM) with a linear kernel. These models offer the advantages of simplicity and interpretability, laying a strong foundation for understanding the basic structure and patterns within our data~\cite{bishop2006pattern}.

On the non-linear front, we employ Quadratic Discriminant Analysis (QDA) and K Nearest Neighbors (KNN). These models are equipped to capture intricate, non-linear relationships in our data, thus offering the opportunity to identify less obvious but potentially significant patterns that linear models may not uncover~\cite{Alpaydin2004IntroductionTM}.

As for tree-based models, we've enlisted a Decision Tree classifier. Decision trees are particularly useful for their straightforward interpretability and ability to handle both categorical and numerical data effectively~\cite{bishop2006pattern, Deisenroth2020MathematicsFM}.

Further, we leverage the power of ensemble decision trees, namely Random Forest and Extra Trees Classifier. These models bring together multiple decision trees to make more robust and accurate predictions. Their combined strength makes them valuable when dealing with complex and imbalanced datasets, as is often the case in anomaly detection~\cite{bishop2006pattern,Nassif2021MachineLF}.

In the boosting decision tree category, we have selected Gradient Boosting, AdaBoost Classifier, XGBoost Classifier, and Light Gradient Boosting Machine. These algorithms iteratively refine their decision-making process to enhance performance, making them well-suited for demanding tasks like anomaly detection~\cite{Nassif2021MachineLF,Ke2017LightGBMAH}.

Moreover, we have included the deep learning model, Multi-Layer Perceptron (MLP), to our repertoire of algorithms. MLP is a type of artificial neural network that consists of multiple layers of interconnected nodes or neurons~\cite{lecun2015deep}. Each neuron applies a non-linear activation function to the weighted sum of its inputs and passes the result to the next layer. MLPs can learn complex relationships between input features and output labels by adjusting the weights and biases during training, using the backpropagation algorithm. This makes them a highly versatile tool for anomaly detection~\cite{Nassif2021MachineLF,Efferen2017AMP}. We have explored different MLP configurations with varying depth and size of hidden linear layers and have run the models on both GPU and CPU for comparison. In alignment with our Green AI goal, we have studied the application of quantization on the MLP model, contrasting full float 32 precision with integer 8 (int8) precision~\cite{Ma2021QuantizationBT}.

Below is a more detailed breakdown of each machine learning model employed in our study, emphasizing their unique contributions to anomaly detection and potential significance within Green AI. The specifics of the experimental setup and individual model configurations will be provided in the forthcoming "Experimental Setup" section~\ref{sec:Exp-setup}.

\subsection{Computational Resources}

As part of our commitment to Green AI principles, we strive to provide transparency and clarity regarding the computational resources used in this study. In addition to advocating energy-efficient AI solutions, these principles also emphasize the importance of reporting the resources necessary to reproduce the results of a study. The information is divided into three categories: Software, Hardware, and Optimization Strategies.

The following software is used in our computational setup:

\begin{itemize}
    \item Operating System: The experiments were conducted on Linux, specifically version 5.15.107+ with an x86\_64 architecture and glibc version 2.31.

    \item Python Version 3.10.11~\cite{python} was the primary programming language used to conduct the research.

    \item Scikit-learn~\cite{scikit-learn}: Version 1.2.2. This Python library was employed to load the algorithms and implement the machine learning models considered in the study.
    
    \item PyCaret~\cite{PyCaret}: Version 3.0.2. This open-source, machine learning library in Python used to compare and train different models from scikit-learn efficiently.

    \item CodeCarbon\footnote{\href{https://mlco2.github.io/codecarbon/}{https://mlco2.github.io/codecarbon/}}: Version 2.2.1.  The Python package used for tracking emissions and energy consumption during model execution adheres to Green AI principles.

    \item PyTorch~\cite{NEURIPS2019_9015} (version 2.0.1) and PyTorch Lightning~\footnote{\href{https://zenodo.org/record/3828935}{https://zenodo.org/record/3828935}} (version 2.0.2), an extension of PyTorch, was employed to promote cleaner and more modular code for the MLP deep learning models, particularly for applying dynamic quantization.

\end{itemize}

The following hardware components are equally important to the operation of our system:

\begin{itemize}
    \item CPU: Our setup was equipped with an Intel(R) Xeon(R) CPU @ 2.20GHz, featuring 2 cores, thereby allowing simultaneous execution of multiple threads or processes.

    \item RAM: We utilized a total of 12GB of RAM, a crucial resource for handling large datasets, training machine learning models, and sustaining multiple tasks or services in memory.

    \item GPU: Our system was outfitted with a Tesla T4 GPU. Given its ability to concurrently manage thousands of threads, the GPU was employed primarily for tasks necessitating high parallelism, such as the training of deep learning models.
\end{itemize}

All machine learning models were executed on the CPU, with the exception of the Multi-Layer Perceptron (MLP). In our study, the MLP was unique in its use of both the CPU and GPU. This utilization of the GPU allowed us to explore its potential for accelerating deep learning tasks, demonstrating the resource efficiencies achievable in this field. Such details underscore Green AI's transparency and provide a comprehensive picture of the resources necessary to replicate our study.

\subsection{Experimental Setup}
\label{sec:Exp-setup}

This section provides a detailed overview of the experimental parameters and procedures employed in our study, encompassing both traditional machine learning models and deep learning counterparts. The objective is to maintain rigorous and comprehensive experimentation, upholding the principles of reproducibility and Green AI.

Our experimental setup includes a variety of machine learning models, as well as the Multi-Layer Perceptron (MLP), a deep learning model. We adhered to the default parameters for each model as specified by the Scikit-learn library, and we set the seed for random state to 1794 to ensure consistency and reproducibility.

Part of our investigation involved studying different configurations of the MLP, exploring both the width and depth of neural networks under the principles of Green AI~\cite{green_ai_best_practices_2022,alzubaidi_review_2021}. The MLP's configuration includes only linear layers with no batch normalization or dropout layers. We used the rectified linear unit (ReLU) as the activation function. The various configurations of MLP we explored are detailed in Tables \ref{table:MLP_CPU_training}, \ref{table:MLP_GPU_training}, \ref{table:MLP_CPU_inference} and \ref{table:MLP_GPU_inference}. For instance, '200,100' refers to two hidden layers with 200 and 100 neurons respectively.

Our assessment matrix for model performance and environmental impact comprised CO${2}$ equivalent (CO${2}$eq) emissions, total energy consumption by CPU, GPU, and RAM (expressed in kW), the F1 Macro score, and the elapsed time during both training and inference stages. This approach ensures that we maintain a balance between resource efficiency and model performance, upholding the principles of Green AI.

To ensure the robustness and reliability of our results, each experiment was conducted five times for each model. Afterwards, we calculated the average values and standard deviations for CO$_{2}$eq emissions, total energy consumption, and elapsed time. These measurements provide insights into the environmental impact of each model and its variability.

The MLP model with best performance was also dynamically quantized using PyTorch and PyTorch Lightning to reduce model size and improve computation speed. Dynamic quantization involves determining the scale factor for activations based on the data range observed at runtime. This method ensures that as much signal as possible from each observed dataset is preserved. The model parameters are converted to int8 form in advance. Arithmetic operations in the quantized model are conducted using vectorized int8 instructions. Accumulation is typically done in int16 or int32 to avoid overflow. These higher precision values are then scaled back to int8 if the next layer is quantized or converted to fp32 for output.

Through detailing these experimental setups and findings, we aim to further the discussion on sustainable and environmentally friendly AI research.

\subsection{Multi-Objective Comparison}

Our study aimed to identify an optimal balance between the performance (measured by F1 Macro score) and the environmental impact (quantified through computation time, CO2 equivalent, and energy consumption) of different machine learning models. To achieve this, we employed a multi-objective optimization approach. This approach demanded a conversion of the F1 Macro score maximization problem into a minimization problem, achieved by considering its reciprocal.

This conversion allowed the application of Pareto optimality principles, which facilitated the computation of the Pareto front - a representation of non-dominated solutions. Each solution on this front represents an optimal trade-off between a model's performance and its environmental impact.

We initially undertook a simplified, two-dimensional optimization problem, focusing on the trade-off between the CO2 equivalent and F1 Macro performance of the models. This approach prioritized the CO2 equivalent as the main environmental metric due to its direct relationship to global warming. Energy consumption, while important, was excluded in this analysis because it varies greatly depending on whether the energy source is renewable or non-renewable.

Similarly, time consumption was not a key priority in our analysis, despite being an essential efficiency metric. This decision aligns with the principles of Green AI, which emphasize reducing environmental impact, even if it means longer computation times.

The two-dimensional optimization offered a clear and simplified view of the relationship between model performance and environmental impact, but we recognized the need for a more comprehensive analysis. Therefore, we conducted a second multi-objective optimization scenario to account for the multi-faceted nature of the problem.

In this more complex scenario, all variables (computation time, energy consumption, CO2 equivalent, and performance) were considered simultaneously. By exploring the interactions of these variables together, we provided a comprehensive understanding of their combined effects and trade-offs in the context of machine learning model optimization.

\subsection{Limitations}
\label{sec:limitations}

Despite our comprehensive analysis, certain limitations of our study should be acknowledged. Some machine learning algorithms values were not explored due to computational constraints. While we aimed to cover a wide range of machine learning algorithms suitable for industrial environments, it is possible that some algorithms were inadvertently overlooked.

Furthermore, in our experimentation, we opted to use default parameter and hyperparameter values instead of performing an exhaustive search for optimal settings. Although this approach provides a baseline for comparison, it may not capture the full potential of each algorithm. Future studies could consider exploring different parameter configurations to further enhance the performance of the algorithms.

Additionally, certain elements such as dropout and batch normalization layers in MLP configurations were omitted from our analysis to prioritize computational efficiency and focus on the impact of depth and width.

Data limitations and inherent imbalances also presented challenges, potentially affecting the precision of our results. Although mitigation strategies were employed, they may not have completely nullified the imbalance effects. These limitations outline potential areas for further research, contributing to the ongoing discussion of Green AI.

\section{Results and Discussion}

The forthcoming section encapsulates an exhaustive examination of diverse machine learning algorithms and configurations. Our investigation seeks to provide an intricate understanding of their performance and environmental footprints, achieved through a multifaceted approach.

The initial subsection \ref{sec:metrics_eval}, offers an in-depth comparative analysis of various evaluation metrics implemented in our study. These metrics, namely Accuracy, Area Under the Curve (AUC), Recall, Precision, F1 score, Kappa statistic, Matthews correlation coefficient (MCC), and F1 Macro, provide a comprehensive perspective on the performance of the assessed machine learning models and the importance of metric selection based on the task at hand related to anomaly detection.

Subsequently, the section \ref{sec:performance_ml} analyzes the environmental consequences associated with the operation of these machine learning models. Our evaluation encompasses factors such as time duration, CO2 equivalent, and energy consumption during the training, cross-validation, and inference stages, thereby shedding light on the sustainability aspects of each model.

The subsequent subsection \ref{sec:perform_mlps} extends this environmental impact analysis to various configurations of Multilayer Perceptrons (MLPs). The discourse elucidates the trade-offs between model complexity and performance while also evaluating their efficiency concerning computational resource requirements.

Finally, the subsection \ref{sec:results_overview} compares the results of traditional machine learning models with MLPs, thereby revealing important insights. This is complemented by a multi-objective comparison which unveils the complex interplay between performance, computational resource consumption, and environmental impact. A discussion of the computed Pareto optimal solutions concludes the section, highlighting the intricate trade-offs in the quest for high-performing yet environmentally sustainable machine learning models. This comprehensive analysis is envisaged to pave the way for future research into environmentally conscious machine learning model optimization.

\subsection{Metrics Evaluation Comparison}
\label{sec:metrics_eval}

In this subsection, we provide a detailed comparison of the test evaluation metrics for the different machine learning algorithms employed in our study. The objective is to assess and contrast the performance of these algorithms based on a range of metrics, including Accuracy, Area Under the Curve (AUC), Recall, Precision, F1 Score, Kappa Statistic, Matthews Correlation Coefficient (MCC) and  F1 Macro.

The table presented in this subsection (Table \ref{table:MLA_metrics_evaluation}) showcases the evaluation results for each algorithm, allowing for a comprehensive analysis of their performance across various metrics.

By examining these metrics, we can gain insights into the strengths and weaknesses of each algorithm, enabling a thorough understanding of their capabilities in addressing the specific task at hand. Furthermore, this comparison enables us to identify algorithms that excel in specific areas and those that provide a more balanced performance across multiple metrics. Our analysis focuses not only on the overall performance metrics but also on the execution time, as it plays a crucial role in real-world applications where efficiency is a critical consideration.

Among the algorithms analyzed, the Random Forest Classifier stands out with exceptional F1 Macro, MCC and Kappa scores, making it proficient at precise predictions and effective class differentiation. It also exhibits high Recall and Precision rates, striking a balance between identifying positive instances and minimizing false positives. However, its longer execution time compared to other algorithms may be a trade-off to consider.

Another powerful algorithm is Extreme Gradient Boosting (XGBoost), which excels in F1 Macro, AUC and MCC scores. Leveraging an ensemble of decision trees and gradient boosting techniques, XGBoost effectively handles complex relationships within the data. This makes it a compelling choice for various classification tasks, particularly when there is a need to deal with imbalanced datasets and produce high-quality predictions.

In contrast, Logistic Regression demonstrates strong performance in Precision and AUC making it suitable for scenarios where linear decision boundaries suffice. However, its effectiveness in capturing non-linear relationships within the data may be limited, potentially affecting its performance in certain situations.

The Extra Trees Classifier also exhibits high Recall and AUC scores. With its robustness against overfitting and ability to handle high-dimensional datasets, it is a viable option for complex classification tasks. Nonetheless, the random feature selection process of the Extra Trees Classifier may compromise interpretability.

Additionally, the Ada Boost Classifier delivers excellent AUC, particularly excelling in handling imbalanced datasets and generating accurate predictions. However, it is important to note that this algorithm may require more computational resources compared to some other alternatives.

\input{tables/table_metrics_evaluation}
\input{tables/table_MLP_metrics_evaluation}

The Decision Tree Classifier, on the other hand, performs well in all scores, owing to its proficiency in capturing complex relationships within the data. It is especially effective when dealing with non-linear decision boundaries. Nevertheless, decision trees are susceptible to overfitting, especially when confronted with noisy or high-dimensional datasets. Regularization techniques and ensemble methods, such as Random Forest, can be employed to mitigate this limitation and enhance the overall performance of the algorithm.

In addition to the machine learning algorithms discussed, we also evaluate the performance of multi-layer perceptrons (MLPs) with various configurations (Table \ref{table:MLP_metrics_evaluation}). MLPs, as neural network-based models, offer a different approach to classification tasks compared to traditional machine learning algorithms. While machine learning algorithms rely on statistical techniques and decision rules, MLPs utilize artificial neural networks with multiple layers of interconnected nodes. These layers allow MLPs to learn and capture intricate patterns and relationships in the data, making them suitable for complex classification problems.

They exhibit strong performance across several common metrics, including Accuracy, Recall, Precision, and F1 score. These metrics indicate the models' ability to effectively capture patterns and make accurate predictions. However, a closer examination of specific metrics reveals variations among the configurations, highlighting their unique strengths and areas of specialization.

MLP\_5 utilizes a single hidden layer with 50 neurons and achieved remarkable AUC and MCC. This configuration demonstrates the ability to effectively capture and learn complex patterns in the data, leading to accurate predictions.

MLP\_7 stands out with its architecture consisting of a single hidden layer with 200 neurons. It showcases excellent AUC and Kappa. This configuration's deeper architecture allows it to capture more intricate relationships within the data, resulting in robust classification performance.

MLP\_4, featuring a deeper architecture with four hidden layers (100, 70, 50, and 20 neurons), demonstrates strong AUC but also has a good performance in Kappa, MCC and F1 Macro. The multi-layer structure enables it to capture intricate data representations, contributing to its classification effectiveness.

When compared to the machine learning algorithms, some MLPs exhibit competitive performance across multiple metrics. Their strong classification capabilities make them viable alternatives for solving complex classification tasks.

However, it is important to consider the specific characteristics of the dataset, computational resources, and interpretability requirements when selecting the most appropriate algorithm or MLP configuration. Depending on the specific context and constraints of the problem, a careful assessment is necessary to determine the optimal choice.

\subsection{Performance and Environmental Impact of Machine Learning Models}
\label{sec:performance_ml}

\input{tables/table_energy}
\input{tables/table_inference}

During the training and cross-validation phases, the Decision Tree and Random Forest Classifiers stood out for their exceptional performance, boasting F1 Macro scores of 0.9101 and 0.9335, respectively. These models completed these phases with high efficiency, minimizing time taken, CO2 emissions, and energy consumption. The Extreme Gradient Boosting model also achieved a notable F1 Macro score of 0.9235, albeit with greater time and energy expenditure.

Contrarily, the K Neighbors Classifier consumed significant resources without delivering high performance, reaching only a 0.6163 F1 Macro score. Naive Bayes and Quadratic Discriminant Analysis models also lagged in performance, failing to offer substantial improvements in time efficiency, CO2 equivalent, or energy consumption.

In the inference and cross-validation phase, the trends generally remained consistent. The Decision Tree and Random Forest Classifiers again excelled in performance and efficiency. The K Neighbors Classifier, however, continued to struggle, consuming the most resources without substantial improvement in the F1 Macro score.

From the findings, a trade-off between performance and environmental impact is apparent. The Decision Tree and Random Forest Classifiers demonstrate the possibility of high performance with reduced environmental impact.

Comparing the single tree Decision Tree Classifier with the ensemble-based Random Forest Classifier reveals the balance between performance and computational cost. The latter, with its enhanced accuracy from multiple decision trees, outperforms the former but at higher computational requirements. Particularly during inference, where speed is often crucial, the Decision Tree model is inherently more efficient due to its simpler structure. Notably, its lower computational demand results in less energy use and CO2 emissions, making it a more eco-friendly choice. Future work can explore enhancing efficiency in high-performing models and improving performance in energy-efficient ones.

\subsection{Performance and Environmental Impact of MLPs}
\label{sec:perform_mlps}

\input{tables/table_MLP_CPU_training}
\input{tables/table_MLP_GPU_training}

During the training and cross-validation phase for MLPs, we observed an interesting balance between model complexity, performance, and resource efficiency. Simpler configurations, such as MLP\_5 with 50 nodes in a single layer, were more time-efficient and consumed less energy across both CPU and GPU platforms. However, more complex configurations, like MLP\_8 with two layers and 300 nodes, delivered higher F1 Macro scores, despite consuming more resources. This points to a clear trade-off between performance and resource efficiency during training.

\input{tables/table_MLP_CPU_Inference}
\input{tables/table_MLP_GPU_Inference}

In the inference phase, similar trends persisted. Complex configurations like MLP\_4 and MLP\_7 exhibited higher F1 Macro scores, albeit at the cost of more time and energy. Conversely, simpler configurations, despite their lower F1 scores, failed to show significant improvement in time efficiency or energy consumption. These findings again highlight the trade-off between performance and complexity in MLPs, pointing towards the need for optimization strategies to balance performance and efficiency.

\input{tables/table_quantization}

When examining the effects of quantization, significant efficiency improvements were noted in the quantized (int8) MLP5 model on CPU, with reduced training time, CO2 emissions, and energy consumption compared to the original (fp32) model. Notably, this did not compromise performance, as F1 Macro scores and validation loss values remained consistent across all configurations. This demonstrates the potential of quantization for enhancing the efficiency of machine learning models without sacrificing performance. Future investigations could further explore the impact of quantization on various MLP configurations and tasks.

\subsection{Comparative Analysis and Key Observations}
\label{sec:results_overview}

The performance comparison between classic machine learning models and MLPs yields some interesting insights. Among the classic machine learning models, the Decision Tree and Random Forest classifiers proved to be the most effective, achieving high F1 Macro scores. Their superiority could be attributed to their inherent ability to handle both linear and non-linear data, making them versatile and efficient across diverse data sets.

On the other hand, the MLPs showed potential for even higher performance, but the results were contingent on the configurations used. More complex configurations, despite being resource-intensive, delivered superior F1 Macro scores, indicating a trade-off between model complexity and performance. It's worth noting that while these configurations required more computational resources, they didn't necessarily compromise on performance, maintaining comparable F1 Macro scores to those of simpler configurations.

In terms of environmental impact, the Decision Tree and Random Forest classifiers were observed to be the most efficient, with low time, CO2 equivalent, and energy consumption values, while maintaining high performance scores. This efficiency could be linked to their inherent simplicity and lower computational complexity compared to other models like the K Neighbors Classifier and Extreme Gradient Boosting.

On the contrary, MLPs generally consumed more time and energy, especially when more complex configurations were used. However, the use of quantization significantly mitigated these effects, maintaining performance while reducing time, CO2 emissions, and energy consumption on the CPU platform. Interestingly, despite leveraging GPU capabilities, the MLPs didn't show a significant reduction in their environmental impact, suggesting an area for future exploration in optimization of MLP configurations for GPU use.

Detailed visualizations of these trade-offs, presented in the appendix \ref{sec:appendix}, provide further insights into how different models balance these factors.

\begin{figure}[t]
\centering
\includegraphics[width=0.75\linewidth]{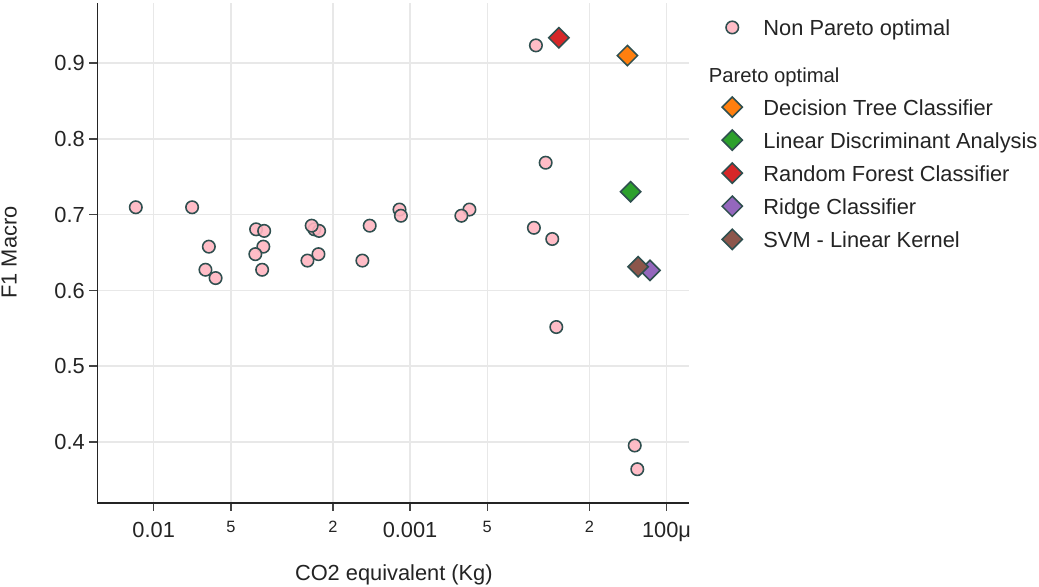}
\caption{Two-dimensional Pareto front demonstrating the trade-off between CO2 equivalent (inverted and log-scaled on the X-axis for improved visualization) and F1 Macro score. Each point on the graph represents an optimal solution considering both environmental impact (CO2 equivalent) and model performance (F1 Macro score).}
\label{fig:2D_Pareto}
\end{figure}

\begin{figure*}[htpb]
\centering
\includegraphics[width=0.75\linewidth]{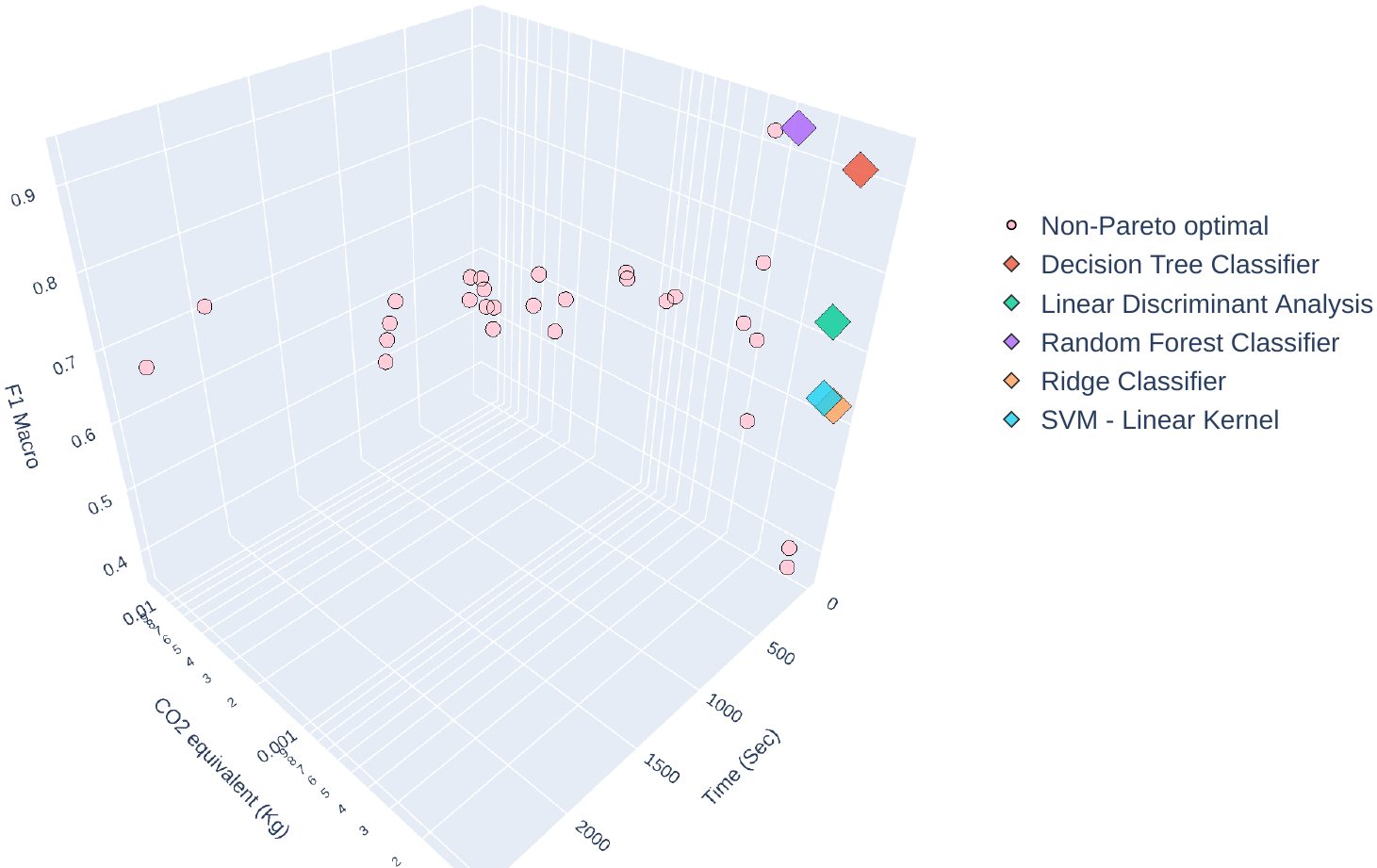}
\caption{Three-dimensional Pareto front showcasing the trade-offs among CO2 equivalent (inverted and log-scaled for improved visualization), F1 Macro score, and computation time. Each point in the 3D space signifies a non-dominated solution considering all three variables simultaneously, providing a comprehensive view of model performance, environmental impact, and computational efficiency.}
\label{fig:3D_pareto_CO2_F1_Time}
\end{figure*}

The Pareto front, computed for both the two-dimensional Figure~\ref{fig:2D_Pareto} and multi-objective scenarios Figure~\ref{fig:3D_pareto_CO2_F1_Time}, revealed that the Decision Tree Classifier, Random Forest Classifier, Ridge Classifier, SVM Linear Kernel, and Linear Discriminant Analysis models presented the best compromise between high performance and low environmental impact. These models, found in the Pareto set, were not dominated by any other model, implying that no other model outperformed them across all objectives simultaneously.

However, the "best" model choice from the Pareto set depends on specific priorities and constraints. For instance, a model that minimizes CO2 emissions might be preferred in contexts prioritizing environmental concerns, even at the cost of slight compromises on performance or longer computation time.

Thus, the Pareto front presents a range of optimal solutions, each aligning with different acceptable trade-offs between performance and environmental impact. Future work could focus on further understanding these trade-offs and exploring strategies to extend the Pareto front, seeking higher performance with lower environmental impact. 



\section{Conclusion}

This research aimed to assess the performance and environmental impact of various machine learning algorithms and Multilayer Perceptrons (MLPs) configurations. Performance was measured through the F1 Macro score, and environmental impact was gauged through the time duration, CO2 equivalent, and energy consumed during training, cross-validation, and inference.

Decision Trees and Random Forests emerged as top performers among the traditional machine learning algorithms with F1 Macro scores of 0.9101 and 0.9335, respectively. Additionally, these models demonstrated exceptional efficiency with regards to time, CO2 equivalent, and energy consumption. On the other hand, the K Neighbors Classifier consumed considerable resources but failed to achieve a high F1 Macro score, demonstrating the inequity between resource consumption and performance. Although Extreme Gradient Boosting achieved a high F1 macro score, it proved to be more resource-intensive than Decision Trees and Random Forests. As evidenced by their lower F1 macro scores, the Naive Bayes and Quadratic Discriminant Analysis models lagged in performance.

In the case of MLPs, we observed a clear trade-off between performance and model complexity. In training and cross-validation, simpler configurations, such as MLP\_5, excelled in terms of time and energy efficiency. Meanwhile, more complex configurations like MLP\_8, despite being resource-intensive, delivered higher F1 Macro scores, underscoring that complexity can enhance performance. Similarly, the inference phase mirrored these findings, with MLP\_4 and MLP\_7 outperforming others on the CPU and GPU, respectively.

According to the Pareto analysis, there is a trade-off between the environmental impact and the performance of the model. It was found that the Decision Tree Classifier, Random Forest Classifier, Ridge Classifier, SVM Linear Kernel, and Linear Discriminant Analysis were the most optimal solutions, striking the best balance between high performance and low environmental impact. It is important to keep in mind that the selection of the 'best' model will be determined by the specific context and priorities, as performance versus environmental concerns are weighted differently.

It should be noted that while traditional algorithms such as Decision Trees and Random Forests have consistently displayed high performance, optimized MLP configurations have demonstrated even greater potential. Nevertheless, these enhanced results came at the cost of increased resource consumption, particularly during training. In contrast, leveraging GPUs over CPUs did not significantly enhance MLPs' performance, suggesting that GPU utilization can be optimized for future studies. The present study illustrates the importance of maintaining a delicate balance between model performance, complexity, and environmental impact, thus shedding light on future considerations for environmentally-conscious machine learning model optimizations.

\section*{Acknowledgments}
The Detecta project is co-financed by the Ministry of Industry, Trade and Tourism of Spain through the line of support for Innovative Business Clusters, in its 2022 call for proposals under the Recovery, Transformation and Resilience Plan.


\bibliographystyle{elsarticle-num}
\bibliography{references.bib} 

\newpage
\appendix

\section{Appendix}
\label{sec:appendix}

\begin{figure}[htpb]
\centering
\includegraphics[width=0.8\linewidth]{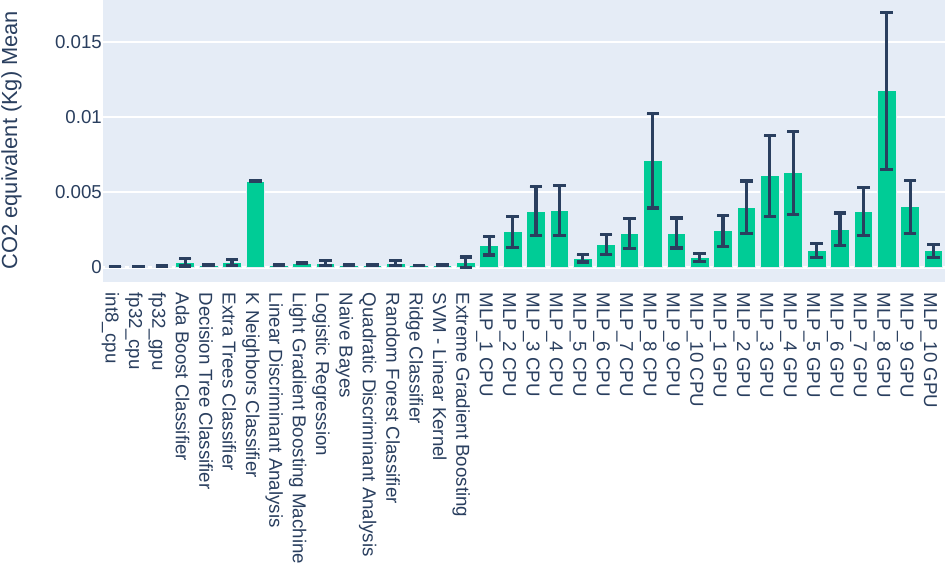}
\caption{Mean CO$_{2}$ equivalent Kg per Model }
\end{figure}

\begin{figure}[htpb]
\centering
\includegraphics[width=0.8\linewidth]{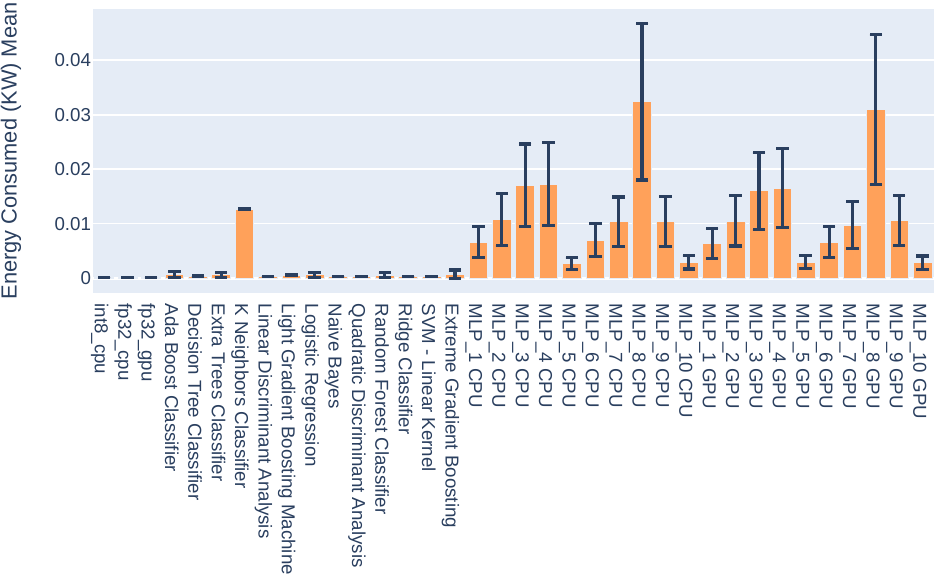}
\caption{Mean energy Consumed (KW) per Model}
\end{figure}

\begin{figure}[htpb]
\centering
\includegraphics[width=0.8\linewidth]{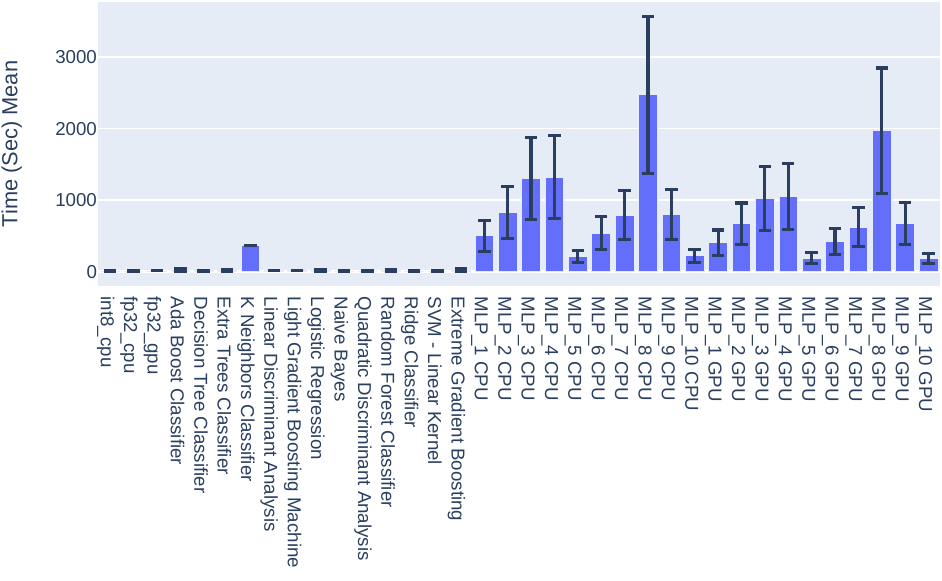}
\caption{Mean Execution Time (sec) per Model}
\end{figure}

\begin{figure}[htpb]
\centering
\includegraphics[width=\linewidth]{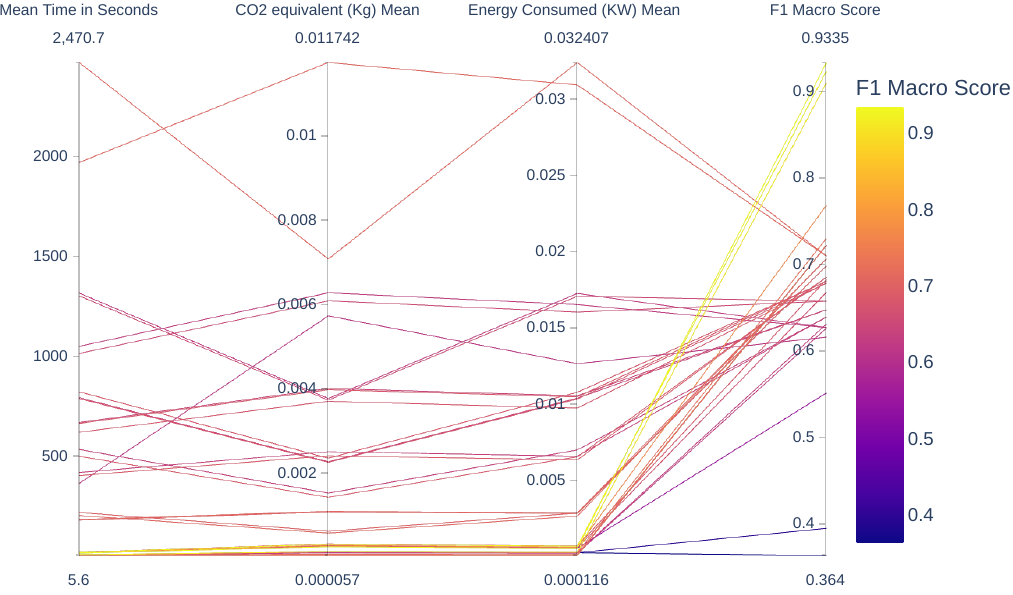}
\caption{Parallel Coordinates Plot showcasing multi-dimensional data analysis. Each vertical line represents a variable, and each point on a line corresponds to a data value for that variable. Lines connecting these points across variables represent individual data instances. This visualization enables the identification of potential relationships and patterns among the multiple variables and metrics in the dataset.}
\end{figure}

\begin{figure}[!htpb]
\captionsetup[subfigure]{labelformat=empty}
\centering
\begin{subfigure}[h]{0.46\linewidth}
\includegraphics[width=\linewidth]{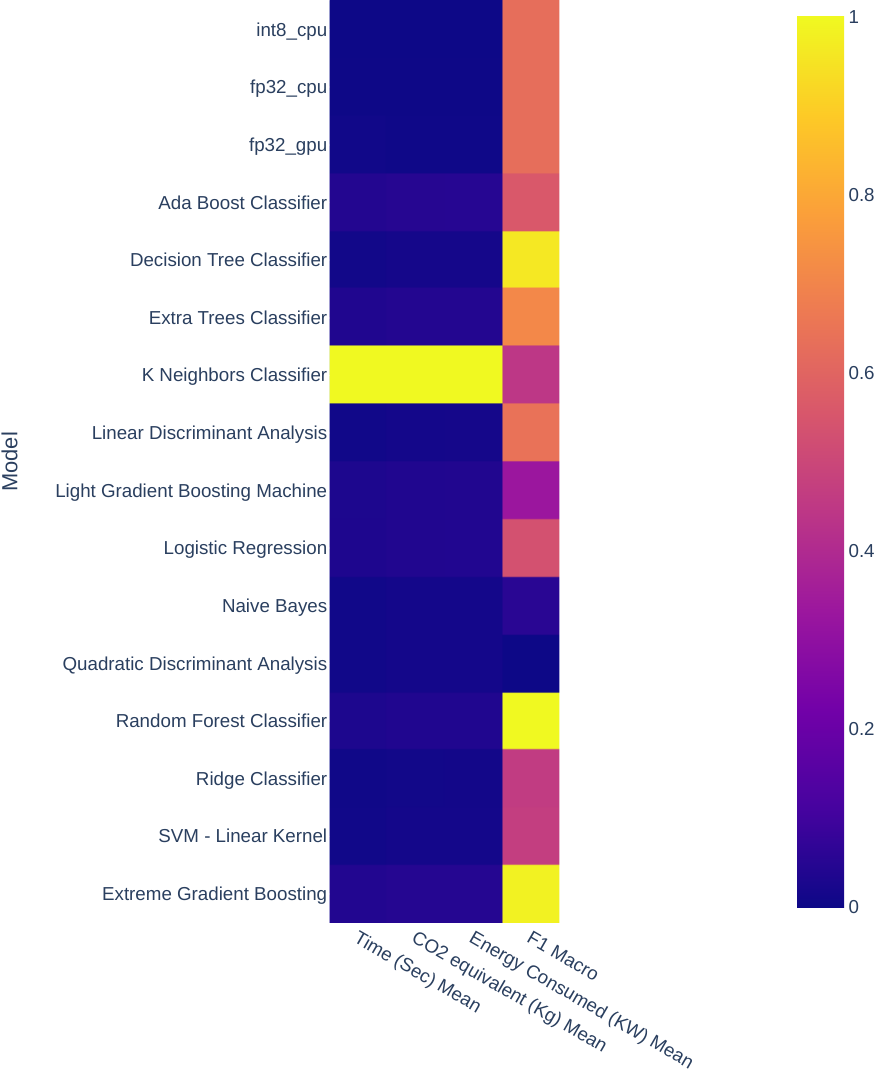}
\caption{Heatmap for Machine Learning models}
\label{real-example-1}
\end{subfigure}
\hspace{0.2cm}
\begin{subfigure}[h]{0.46\linewidth}
\includegraphics[width=\linewidth]{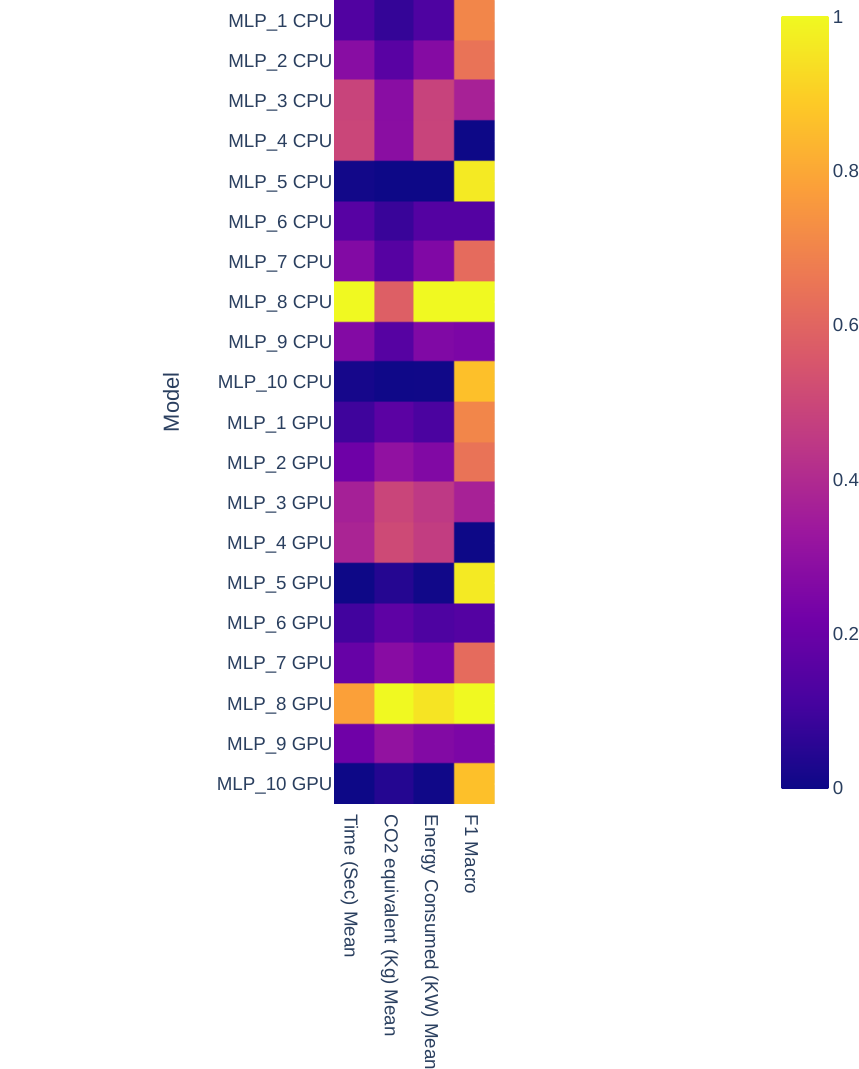}
\caption{Heatmap for MLPs}
\label{real-example-2}
\end{subfigure}
\end{figure}

\end{document}

%% file: tables/table_metrics_evaluation.tex
\begin{table}[htbp]
\centering
\caption{Machine Learning Algorithms Metrics Evaluation}
\label{table:MLA_metrics_evaluation}
\resizebox{\linewidth}{!}{%
\begin{tblr}{
  row{1} = {c},
  column{9} = {c},
  vline{2-9} = {2-14}{},
  hline{1-2,15} = {-}{},
}
\textbf{Model}                  & \textbf{Accuracy} & \textbf{AUC} & \textbf{Recall} & \textbf{Prec.} & \textbf{F1} & \textbf{Kappa} & \textbf{MCC} & \textbf{F1 Macro} \\
Ada Boost Classifier            & 0.9994            & 0.9869       & 0.9994          & 0.9994         & 0.9994      & 0.7473         & 0.7667       & 0.6537            \\
Decision Tree Classifier        & 0.9994            & 0.857        & 0.9994          & 0.9994         & 0.9994      & 0.7657         & 0.7677       & 0.899             \\
Extra Trees Classifier          & 0.9995            & 0.943        & 0.9995          & 0.9995         & 0.9994      & 0.7705         & 0.7916       & 0.7103            \\
K Neighbors Classifier          & 0.9994            & 0.8529       & 0.9994          & 0.9993         & 0.9993      & 0.7423         & 0.7662       & 0.595             \\
Light Gradient Boosting Machine & 0.9984            & 0.7741       & 0.9984          & 0.9986         & 0.9985      & 0.464          & 0.4649       & 0.5907            \\
Linear Discriminant Analysis    & 0.9991            & 0.8373       & 0.9991          & 0.9995         & 0.9992      & 0.673          & 0.673        & 0.695             \\
Logistic Regression             & 0.9995            & 0.9956       & 0.9995          & 0.9994         & 0.9994      & 0.7607         & 0.7816       & 0.6548            \\
Naive Bayes                     & 0.969             & 0.9946       & 0.969           & 0.9987         & 0.9832      & 0.0702         & 0.1796       & 0.3904            \\
Quadratic Discriminant Analysis & 0.9794            & 0.9647       & 0.9794          & 0.9987         & 0.9884      & 0.1001         & 0.2142       & 0.4734            \\
Random Forest Classifier        & 0.9996            & 0.9391       & 0.9996          & 0.9996         & 0.9995      & 0.8073         & 0.8212       & 0.9137            \\
Ridge Classifier                & 0.9994            & 0            & 0.9994          & 0.9993         & 0.9993      & 0.746          & 0.7713       & 0.595             \\
SVM - Linear Kernel             & 0.9993            & 0            & 0.9993          & 0.9992         & 0.9992      & 0.6734         & 0.7125       & 0.5703            \\
Extreme Gradient Boosting       & 0.9995            & 0.9961       & 0.9995          & 0.9995         & 0.9995      & 0.8035         & 0.8165       & 0.8949            
\end{tblr}
}
\end{table}

%% file: tables/table_MLP_metrics_evaluation.tex
\begin{table}[htbp]
\centering
\caption{MLP Metrics Evaluation Comparison}
\label{table:MLP_metrics_evaluation}
\resizebox{\linewidth}{!}{%
\begin{tblr}{
  column{3} = {c},
  column{11} = {c},
  cell{2}{4} = {c},
  cell{2}{5} = {c},
  cell{2}{6} = {c},
  cell{3}{4} = {c},
  cell{3}{5} = {c},
  cell{3}{6} = {c},
  cell{4}{4} = {c},
  cell{4}{5} = {c},
  cell{4}{6} = {c},
  cell{5}{4} = {c},
  cell{5}{5} = {c},
  cell{5}{6} = {c},
  cell{6}{4} = {c},
  cell{6}{5} = {c},
  cell{6}{6} = {c},
  cell{7}{4} = {c},
  cell{7}{5} = {c},
  cell{7}{6} = {c},
  cell{8}{4} = {c},
  cell{8}{5} = {c},
  cell{8}{6} = {c},
  cell{9}{4} = {c},
  cell{9}{5} = {c},
  cell{9}{6} = {c},
  cell{10}{4} = {c},
  cell{10}{5} = {c},
  cell{10}{6} = {c},
  cell{11}{4} = {c},
  cell{11}{5} = {c},
  cell{11}{6} = {c},
  vline{4-11} = {2-11}{},
  hline{1-2,12} = {-}{},
}
\textbf{Model} & \textbf{Config} & \textbf{Parameters} & \textbf{Accuracy} & \textbf{AUC} & \textbf{Recall} & \textbf{Prec.} & \textbf{F1} & \textbf{Kappa} & \textbf{MCC} & \textbf{F1 Macro} \\
MLP\_1         & 100             & 2903                & 0.9995            & 0.9936       & 0.9995          & 0.9995         & 0.9994      & 0.7642         & 0.7819       & 0.6578            \\
MLP\_2         & 100, 70         & 9883                & 0.9995            & 0            & 0.9995          & 0.9994         & 0.9994      & 0.7584         & 0.7815       & 0.5992            \\
MLP\_3         & 100, 70, 50     & 13373               & 0.9995            & 0            & 0.9995          & 0.9994         & 0.9994      & 0.7584         & 0.7815       & 0.5992            \\
MLP\_4         & 100, 70, 50, 20 & 14303               & 0.9994            & 0.9922       & 0.9994          & 0.9994         & 0.9994      & 0.7509         & 0.7639       & 0.7031            \\
MLP\_5         & 50              & 1453                & 0.9995            & 0.9946       & 0.9995          & 0.9995         & 0.9994      & 0.7824         & 0.8016       & 0.7144            \\
MLP\_6         & 50, 50          & 4003                & 0.9995            & 0            & 0.9995          & 0.9994         & 0.9994      & 0.7584         & 0.7815       & 0.5992            \\
MLP\_7         & 200             & 5803                & 0.9995            & 0.9874       & 0.9995          & 0.9995         & 0.9994      & 0.7832         & 0.7936       & 0.7143            \\
MLP\_8         & 200, 100        & 25603               & 0.9995            & 0            & 0.9995          & 0.9993         & 0.9994      & 0.7547         & 0.7765       & 0.5978            \\
MLP\_9         & 50, 40, 30, 20  & 5253                & 0.9995            & 0.995        & 0.9995          & 0.9995         & 0.9994      & 0.7645         & 0.7866       & 0.6598            \\
MLP\_10        & 40, 10          & 1483                & 0.9995            & 0            & 0.9995          & 0.9994         & 0.9994      & 0.7584         & 0.7815       & 0.5992            
\end{tblr}
}
\end{table}

%% file: tables/table_energy.tex
\begin{table}[htbp]
\centering
\caption{Duration, Emission, Energy and Performance in Training and Cross Validation phase for Machine Learning algorithms}
\resizebox{\linewidth}{!}{%
\begin{tblr}{
  row{1} = {c},
  row{2} = {c},
  cell{1}{1} = {r=2}{},
  cell{1}{2} = {c=2}{},
  cell{1}{4} = {c=2}{},
  cell{1}{6} = {c=2}{},
  cell{1}{8} = {r=2}{},
  cell{3}{2} = {c},
  cell{3}{3} = {c},
  cell{3}{4} = {c},
  cell{3}{5} = {c},
  cell{3}{6} = {c},
  cell{3}{7} = {c},
  cell{3}{8} = {c},
  cell{4}{2} = {c},
  cell{4}{3} = {c},
  cell{4}{4} = {c},
  cell{4}{5} = {c},
  cell{4}{6} = {c},
  cell{4}{7} = {c},
  cell{4}{8} = {c},
  cell{5}{2} = {c},
  cell{5}{3} = {c},
  cell{5}{4} = {c},
  cell{5}{5} = {c},
  cell{5}{6} = {c},
  cell{5}{7} = {c},
  cell{5}{8} = {c},
  cell{6}{2} = {c},
  cell{6}{3} = {c},
  cell{6}{4} = {c},
  cell{6}{5} = {c},
  cell{6}{6} = {c},
  cell{6}{7} = {c},
  cell{6}{8} = {c},
  cell{7}{2} = {c},
  cell{7}{3} = {c},
  cell{7}{4} = {c},
  cell{7}{5} = {c},
  cell{7}{6} = {c},
  cell{7}{7} = {c},
  cell{7}{8} = {c},
  cell{8}{2} = {c},
  cell{8}{3} = {c},
  cell{8}{4} = {c},
  cell{8}{5} = {c},
  cell{8}{6} = {c},
  cell{8}{7} = {c},
  cell{8}{8} = {c},
  cell{9}{2} = {c},
  cell{9}{3} = {c},
  cell{9}{4} = {c},
  cell{9}{5} = {c},
  cell{9}{6} = {c},
  cell{9}{7} = {c},
  cell{9}{8} = {c},
  cell{10}{2} = {c},
  cell{10}{3} = {c},
  cell{10}{4} = {c},
  cell{10}{5} = {c},
  cell{10}{6} = {c},
  cell{10}{7} = {c},
  cell{10}{8} = {c},
  cell{11}{2} = {c},
  cell{11}{3} = {c},
  cell{11}{4} = {c},
  cell{11}{5} = {c},
  cell{11}{6} = {c},
  cell{11}{7} = {c},
  cell{11}{8} = {c},
  cell{12}{2} = {c},
  cell{12}{3} = {c},
  cell{12}{4} = {c},
  cell{12}{5} = {c},
  cell{12}{6} = {c},
  cell{12}{7} = {c},
  cell{12}{8} = {c},
  cell{13}{2} = {c},
  cell{13}{3} = {c},
  cell{13}{4} = {c},
  cell{13}{5} = {c},
  cell{13}{6} = {c},
  cell{13}{7} = {c},
  cell{13}{8} = {c},
  cell{14}{2} = {c},
  cell{14}{3} = {c},
  cell{14}{4} = {c},
  cell{14}{5} = {c},
  cell{14}{6} = {c},
  cell{14}{7} = {c},
  cell{14}{8} = {c},
  cell{15}{2} = {c},
  cell{15}{3} = {c},
  cell{15}{4} = {c},
  cell{15}{5} = {c},
  cell{15}{6} = {c},
  cell{15}{7} = {c},
  cell{15}{8} = {c},
  vline{2,4,6,8} = {3-15}{},
  hline{1,16} = {-}{0.08em},
  hline{3} = {-}{},
}
\textbf{ Model}                 & \textbf{Time (Sec) } &              & \textbf{CO2 equivalent (Kg) } &              & \textbf{Energy Consumed (KW) } &              & \textbf{F1 Macro} \\
                                & \textbf{Mean}        & \textbf{Std} & \textbf{Mean}                 & \textbf{Std} & \textbf{Mean}                  & \textbf{Std} &                   \\
Ada Boost Classifier            & 20.98                & 15.91        & 3.29E-04                      & 2.49E-04     & 7.27E-04                       & 5.51E-04     & 0.6827            \\
Decision Tree Classifier        & 9.03                 & 2.16         & 1.42E-04                      & 3.40E-05     & 3.13E-04                       & 7.50E-05     & 0.9101            \\
Extra Trees Classifier          & 18.85                & 13.13        & 2.96E-04                      & 2.06E-04     & 6.53E-04                       & 4.55E-04     & 0.7686            \\
K Neighbors Classifier          & 365.78               & 3.78         & 5.73E-03                      & 5.90E-05     & 1.27E-02                       & 1.31E-04     & 0.6163            \\
Linear Discriminant Analysis    & 8.79                 & 0.65         & 1.38E-04                      & 1.00E-05     & 3.05E-04                       & 2.30E-05     & 0.7303            \\
Light Gradient Boosting Machine & 17.15                & 2.80         & 2.69E-04                      & 4.40E-05     & 5.94E-04                       & 9.70E-05     & 0.5517            \\
Logistic Regression             & 17.79                & 12.42        & 2.79E-04                      & 1.95E-04     & 6.16E-04                       & 4.30E-04     & 0.6679            \\
Naive Bayes                     & 8.50                 & 1.04         & 1.33E-04                      & 1.60E-05     & 2.94E-04                       & 3.60E-05     & 0.3953            \\
Quadratic Discriminant Analysis & 8.31                 & 1.05         & 1.30E-04                      & 1.60E-05     & 2.88E-04                       & 3.60E-05     & 0.364             \\
Random Forest Classifier        & 16.78                & 11.29        & 2.63E-04                      & 1.77E-04     & 5.81E-04                       & 3.91E-04     & \textbf{0.9335}            \\
Ridge Classifier                & 7.41                 & 0.64         & 1.16E-04                      & 1.00E-05     & 2.57E-04                       & 2.20E-05     & 0.6265            \\
SVM - Linear Kernel             & 8.22                 & 1.33         & 1.29E-04                      & 2.10E-05     & 2.85E-04                       & 4.60E-05     & 0.6313            \\
Extreme Gradient Boosting       & 20.62                & 22.75        & 3.23E-04                      & 3.57E-04     & 7.14E-04                       & 7.88E-04     & 0.9235            
\end{tblr}
}
\end{table}

%% file: tables/table_inference.tex
\begin{table}[htbp]
\centering
\caption{Duration, Emission, Energy and Performance in Inference phase for Machine Learning algorithms}
\resizebox{\linewidth}{!}{%
\begin{tblr}{
  row{1} = {c},
  row{2} = {c},
  cell{1}{1} = {r=2}{},
  cell{1}{2} = {c=2}{},
  cell{1}{4} = {c=2}{},
  cell{1}{6} = {c=2}{},
  cell{1}{8} = {r=2}{},
  cell{3}{2} = {c},
  cell{3}{3} = {c},
  cell{3}{4} = {c},
  cell{3}{5} = {c},
  cell{3}{6} = {c},
  cell{3}{7} = {c},
  cell{3}{8} = {c},
  cell{4}{2} = {c},
  cell{4}{3} = {c},
  cell{4}{4} = {c},
  cell{4}{5} = {c},
  cell{4}{6} = {c},
  cell{4}{7} = {c},
  cell{4}{8} = {c},
  cell{5}{2} = {c},
  cell{5}{3} = {c},
  cell{5}{4} = {c},
  cell{5}{5} = {c},
  cell{5}{6} = {c},
  cell{5}{7} = {c},
  cell{5}{8} = {c},
  cell{6}{2} = {c},
  cell{6}{3} = {c},
  cell{6}{4} = {c},
  cell{6}{5} = {c},
  cell{6}{6} = {c},
  cell{6}{7} = {c},
  cell{6}{8} = {c},
  cell{7}{2} = {c},
  cell{7}{3} = {c},
  cell{7}{4} = {c},
  cell{7}{5} = {c},
  cell{7}{6} = {c},
  cell{7}{7} = {c},
  cell{7}{8} = {c},
  cell{8}{2} = {c},
  cell{8}{3} = {c},
  cell{8}{4} = {c},
  cell{8}{5} = {c},
  cell{8}{6} = {c},
  cell{8}{7} = {c},
  cell{8}{8} = {c},
  cell{9}{2} = {c},
  cell{9}{3} = {c},
  cell{9}{4} = {c},
  cell{9}{5} = {c},
  cell{9}{6} = {c},
  cell{9}{7} = {c},
  cell{9}{8} = {c},
  cell{10}{2} = {c},
  cell{10}{3} = {c},
  cell{10}{4} = {c},
  cell{10}{5} = {c},
  cell{10}{6} = {c},
  cell{10}{7} = {c},
  cell{10}{8} = {c},
  cell{11}{2} = {c},
  cell{11}{3} = {c},
  cell{11}{4} = {c},
  cell{11}{5} = {c},
  cell{11}{6} = {c},
  cell{11}{7} = {c},
  cell{11}{8} = {c},
  cell{12}{2} = {c},
  cell{12}{3} = {c},
  cell{12}{4} = {c},
  cell{12}{5} = {c},
  cell{12}{6} = {c},
  cell{12}{7} = {c},
  cell{12}{8} = {c},
  cell{13}{2} = {c},
  cell{13}{3} = {c},
  cell{13}{4} = {c},
  cell{13}{5} = {c},
  cell{13}{6} = {c},
  cell{13}{7} = {c},
  cell{13}{8} = {c},
  cell{14}{2} = {c},
  cell{14}{3} = {c},
  cell{14}{4} = {c},
  cell{14}{5} = {c},
  cell{14}{6} = {c},
  cell{14}{7} = {c},
  cell{14}{8} = {c},
  cell{15}{2} = {c},
  cell{15}{3} = {c},
  cell{15}{4} = {c},
  cell{15}{5} = {c},
  cell{15}{6} = {c},
  cell{15}{7} = {c},
  cell{15}{8} = {c},
  vline{2,4,6,8} = {3-15}{},
  hline{1,16} = {-}{0.08em},
  hline{3} = {-}{},
}
\textbf{ Model}                 & \textbf{Time (Sec)} &              & \textbf{CO2 equivalent (Kg)} &              & \textbf{Energy Consumed (KW)} &              & \textbf{F1 Macro } \\
                                & \textbf{Mean}       & \textbf{Std} & \textbf{Mean}                & \textbf{Std} & \textbf{Mean}                 & \textbf{Std} &                    \\
Ada Boost Classifier            & 3.80                & 0.33         & 6.00E-05                     & 5.10E-06     & 1.32E-04                      & 1.13E-05     & 0.6537             \\
Decision Tree Classifier        & 2.48                & 0.02         & 3.90E-05                     & 3.44E-07     & 8.60E-05                      & 7.59E-07     & 0.899              \\
Extra Trees Classifier          & 3.76                & 0.66         & 5.90E-05                     & 1.04E-05     & 1.30E-04                      & 2.30E-05     & 0.7103             \\
K Neighbors Classifier          & 172.78              & 0.34         & 2.71E-03                     & 5.12E-06     & 5.98E-03                      & 1.13E-05     & 0.595              \\
Linear Discriminant Analysis    & 2.73                & 0.23         & 4.30E-05                     & 3.61E-06     & 9.50E-05                      & 7.97E-06     & 0.695              \\
Light Gradient Boosting Machine & 4.98                & 0.49         & 7.80E-05                     & 7.74E-06     & 1.73E-04                      & 1.71E-05     & 0.5907             \\
Logistic Regression             & 2.54                & 0.01         & 4.00E-05                     & 2.51E-07     & 8.80E-05                      & 5.54E-07     & 0.6548             \\
Naive Bayes                     & 2.54                & 0.09         & 4.00E-05                     & 1.43E-06     & 8.80E-05                      & 3.16E-06     & 0.3904             \\
Quadratic Discriminant Analysis & 2.62                & 0.03         & 4.10E-05                     & 4.87E-07     & 9.10E-05                      & 1.08E-06     & 0.4734             \\
Random Forest Classifier        & 3.53                & 0.42         & 5.50E-05                     & 6.64E-06     & 1.22E-04                      & 1.47E-05     & \textbf{0.9137}    \\
Ridge Classifier                & 2.60                & 0.42         & 4.10E-05                     & 6.54E-06     & 9.00E-05                      & 1.44E-05     & 0.595              \\
SVM - Linear Kernel             & 2.41                & 0.01         & 3.80E-05                     & 1.57E-07     & 8.30E-05                      & 3.46E-07     & 0.5703             \\
Extreme Gradient Boosting       & 2.86                & 0.03         & 4.50E-05                     & 4.98E-07     & 9.90E-05                      & 1.10E-06     & 0.8949             
\end{tblr}
}
\end{table}

%% file: tables/table_MLP_CPU_training.tex
\begin{table}[htbp]
\centering
\caption{Duration, Emission, Energy and Performance in Training and Cross Validation phase for Multi Perceptron Model deep learning model with CPU}
\label{table:MLP_CPU_training}
\resizebox{\linewidth}{!}{%
\begin{tblr}{
  row{1} = {c},
  row{2} = {c},
  cell{1}{1} = {r=2}{},
  cell{1}{2} = {r=2}{},
  cell{1}{3} = {r=2}{},
  cell{1}{4} = {c=2}{},
  cell{1}{6} = {c=2}{},
  cell{1}{8} = {c=2}{},
  cell{1}{10} = {r=2}{},
  cell{3}{3} = {c},
  cell{3}{4} = {c},
  cell{3}{5} = {c},
  cell{3}{6} = {c},
  cell{3}{7} = {c},
  cell{3}{8} = {c},
  cell{3}{9} = {c},
  cell{3}{10} = {c},
  cell{4}{3} = {c},
  cell{4}{4} = {c},
  cell{4}{5} = {c},
  cell{4}{6} = {c},
  cell{4}{7} = {c},
  cell{4}{8} = {c},
  cell{4}{9} = {c},
  cell{4}{10} = {c},
  cell{5}{3} = {c},
  cell{5}{4} = {c},
  cell{5}{5} = {c},
  cell{5}{6} = {c},
  cell{5}{7} = {c},
  cell{5}{8} = {c},
  cell{5}{9} = {c},
  cell{5}{10} = {c},
  cell{6}{3} = {c},
  cell{6}{4} = {c},
  cell{6}{5} = {c},
  cell{6}{6} = {c},
  cell{6}{7} = {c},
  cell{6}{8} = {c},
  cell{6}{9} = {c},
  cell{6}{10} = {c},
  cell{7}{3} = {c},
  cell{7}{4} = {c},
  cell{7}{5} = {c},
  cell{7}{6} = {c},
  cell{7}{7} = {c},
  cell{7}{8} = {c},
  cell{7}{9} = {c},
  cell{7}{10} = {c},
  cell{8}{3} = {c},
  cell{8}{4} = {c},
  cell{8}{5} = {c},
  cell{8}{6} = {c},
  cell{8}{7} = {c},
  cell{8}{8} = {c},
  cell{8}{9} = {c},
  cell{8}{10} = {c},
  cell{9}{3} = {c},
  cell{9}{4} = {c},
  cell{9}{5} = {c},
  cell{9}{6} = {c},
  cell{9}{7} = {c},
  cell{9}{8} = {c},
  cell{9}{9} = {c},
  cell{9}{10} = {c},
  cell{10}{3} = {c},
  cell{10}{4} = {c},
  cell{10}{5} = {c},
  cell{10}{6} = {c},
  cell{10}{7} = {c},
  cell{10}{8} = {c},
  cell{10}{9} = {c},
  cell{10}{10} = {c},
  cell{11}{3} = {c},
  cell{11}{4} = {c},
  cell{11}{5} = {c},
  cell{11}{6} = {c},
  cell{11}{7} = {c},
  cell{11}{8} = {c},
  cell{11}{9} = {c},
  cell{11}{10} = {c},
  cell{12}{3} = {c},
  cell{12}{4} = {c},
  cell{12}{5} = {c},
  cell{12}{6} = {c},
  cell{12}{7} = {c},
  cell{12}{8} = {c},
  cell{12}{9} = {c},
  cell{12}{10} = {c},
  vline{6,8,10} = {3-12}{},
  hline{1,13} = {-}{0.08em},
  hline{3} = {-}{},
}
\textbf{Model} & \textbf{Config} & \textbf{Parameters} & \textbf{Time (Sec)} &              & \textbf{CO2 equivalent (Kg)} &              & \textbf{Energy Consumed (KW)} &              & \textbf{F1 Macro} \\
               &                 &                     & \textbf{Mean}       & \textbf{Std} & \textbf{Mean}                & \textbf{Std} & \textbf{Mean}                 & \textbf{Std} &                   \\
MLP\_1         & 100             & 2903                & 501.15              & 217.05       & 1.44E-03                     & 6.22E-04     & 6.57E-03                      & 2.85E-03     & 0.6855            \\
MLP\_2         & 100, 70         & 9883                & 824.03              & 361.06       & 2.36E-03                     & 1.03E-03     & 1.08E-02                      & 4.74E-03     & 0.6807            \\
MLP\_3         & 100, 70, 50     & 13373               & 1303.34             & 574.55       & 3.74E-03                     & 1.65E-03     & 1.71E-02                      & 7.54E-03     & 0.6578            \\
MLP\_4         & 100, 70, 50, 20 & 14303               & 1317.20             & 581.43       & 3.78E-03                     & 1.67E-03     & 1.73E-02                      & 7.63E-03     & 0.6273            \\
MLP\_5         & 50              & 1453                & 204.99              & 84.56        & 5.87E-04                     & 2.42E-04     & 2.69E-03                      & 1.11E-03     & 0.7068            \\
MLP\_6         & 50, 50          & 4003                & 535.39              & 232.56       & 1.53E-03                     & 6.67E-04     & 7.02E-03                      & 3.05E-03     & 0.6394            \\
MLP\_7         & 200             & 5803                & 789.98              & 346.03       & 2.26E-03                     & 9.92E-04     & 1.04E-02                      & 4.54E-03     & 0.6786            \\
MLP\_8         & 200, 100        & 25603               & 2470.65             & 1097.28      & 7.08E-03                     & 3.14E-03     & 3.24E-02                      & 1.44E-02     & 0.7098            \\
MLP\_9         & 50, 40, 30, 20  & 5253                & 794.30              & 348.35       & 2.28E-03                     & 9.98E-04     & 1.04E-02                      & 4.57E-03     & 0.6479            \\
MLP\_10        & 40, 10          & 1483                & 220.22              & 91.62        & 6.31E-04                     & 2.63E-04     & 2.89E-03                      & 1.20E-03     & 0.6986            
\end{tblr}
}
\end{table}

%% file: tables/table_MLP_GPU_training.tex
\begin{table}[htbp]
\centering
\caption{Duration, Emission, Energy and Performance in Training and Cross Validation phase for Multi Perceptron Model deep learning model with GPU}
\label{table:MLP_GPU_training}
\resizebox{\linewidth}{!}{%
\begin{tblr}{
  row{1} = {c},
  row{2} = {c},
  cell{1}{1} = {r=2}{},
  cell{1}{2} = {r=2}{},
  cell{1}{3} = {r=2}{},
  cell{1}{4} = {c=2}{},
  cell{1}{6} = {c=2}{},
  cell{1}{8} = {c=2}{},
  cell{1}{10} = {r=2}{},
  cell{3}{3} = {c},
  cell{3}{4} = {c},
  cell{3}{5} = {c},
  cell{3}{6} = {c},
  cell{3}{7} = {c},
  cell{3}{8} = {c},
  cell{3}{9} = {c},
  cell{3}{10} = {c},
  cell{4}{3} = {c},
  cell{4}{4} = {c},
  cell{4}{5} = {c},
  cell{4}{6} = {c},
  cell{4}{7} = {c},
  cell{4}{8} = {c},
  cell{4}{9} = {c},
  cell{4}{10} = {c},
  cell{5}{3} = {c},
  cell{5}{4} = {c},
  cell{5}{5} = {c},
  cell{5}{6} = {c},
  cell{5}{7} = {c},
  cell{5}{8} = {c},
  cell{5}{9} = {c},
  cell{5}{10} = {c},
  cell{6}{3} = {c},
  cell{6}{4} = {c},
  cell{6}{5} = {c},
  cell{6}{6} = {c},
  cell{6}{7} = {c},
  cell{6}{8} = {c},
  cell{6}{9} = {c},
  cell{6}{10} = {c},
  cell{7}{3} = {c},
  cell{7}{4} = {c},
  cell{7}{5} = {c},
  cell{7}{6} = {c},
  cell{7}{7} = {c},
  cell{7}{8} = {c},
  cell{7}{9} = {c},
  cell{7}{10} = {c},
  cell{8}{3} = {c},
  cell{8}{4} = {c},
  cell{8}{5} = {c},
  cell{8}{6} = {c},
  cell{8}{7} = {c},
  cell{8}{8} = {c},
  cell{8}{9} = {c},
  cell{8}{10} = {c},
  cell{9}{3} = {c},
  cell{9}{4} = {c},
  cell{9}{5} = {c},
  cell{9}{6} = {c},
  cell{9}{7} = {c},
  cell{9}{8} = {c},
  cell{9}{9} = {c},
  cell{9}{10} = {c},
  cell{10}{3} = {c},
  cell{10}{4} = {c},
  cell{10}{5} = {c},
  cell{10}{6} = {c},
  cell{10}{7} = {c},
  cell{10}{8} = {c},
  cell{10}{9} = {c},
  cell{10}{10} = {c},
  cell{11}{3} = {c},
  cell{11}{4} = {c},
  cell{11}{5} = {c},
  cell{11}{6} = {c},
  cell{11}{7} = {c},
  cell{11}{8} = {c},
  cell{11}{9} = {c},
  cell{11}{10} = {c},
  cell{12}{3} = {c},
  cell{12}{4} = {c},
  cell{12}{5} = {c},
  cell{12}{6} = {c},
  cell{12}{7} = {c},
  cell{12}{8} = {c},
  cell{12}{9} = {c},
  cell{12}{10} = {c},
  vline{4,6,8,10} = {3-12}{},
  hline{1,13} = {-}{0.08em},
  hline{3} = {-}{},
}
\textbf{Model} & \textbf{Config} & \textbf{Parameters} & \textbf{Time (Sec)} &              & \textbf{CO2 equivalent (Kg)} &              & \textbf{Energy Consumed (KW)} &              & \textbf{F1 Macro} \\
               &                 &                     & \textbf{Mean}       & \textbf{Std} & \textbf{Mean}                & \textbf{Std} & \textbf{Mean}                 & \textbf{Std} &                   \\
MLP\_1         & 100             & 2903                & 404.78              & 175.32       & 2.42E-03                     & 1.05E-03     & 6.37E-03                      & 2.76E-03     & 0.6855            \\
MLP\_2         & 100, 70         & 9883                & 665.94              & 291.95       & 3.99E-03                     & 1.75E-03     & 1.05E-02                      & 4.60E-03     & 0.6807            \\
MLP\_3         & 100, 70, 50     & 13373               & 1015.74             & 448.64       & 6.09E-03                     & 2.69E-03     & 1.61E-02                      & 7.09E-03     & 0.6578            \\
MLP\_4         & 100, 70, 50, 20 & 14303               & 1051.51             & 464.61       & 6.28E-03                     & 2.78E-03     & 1.65E-02                      & 7.31E-03     & 0.6273            \\
MLP\_5         & 50              & 1453                & 184.74              & 76.64        & 1.10E-03                     & 4.56E-04     & 2.90E-03                      & 1.20E-03     & 0.7068            \\
MLP\_6         & 50, 50          & 4003                & 420.28              & 182.31       & 2.51E-03                     & 1.09E-03     & 6.62E-03                      & 2.87E-03     & 0.6394            \\
MLP\_7         & 200             & 5803                & 621.77              & 272.29       & 3.71E-03                     & 1.62E-03     & 9.77E-03                      & 4.28E-03     & 0.6786            \\
MLP\_8         & 200, 100        & 25603               & 1970.52             & 875.20       & 1.17E-02                     & 5.22E-03     & 3.09E-02                      & 1.37E-02     & 0.7098            \\
MLP\_9         & 50, 40, 30, 20  & 5253                & 670.54              & 294.59       & 4.01E-03                     & 1.76E-03     & 1.06E-02                      & 4.65E-03     & 0.6479            \\
MLP\_10        & 40, 10          & 1483                & 181.83              & 76.03        & 1.09E-03                     & 4.54E-04     & 2.86E-03                      & 1.20E-03     & 0.6986            
\end{tblr}
}
\end{table}

%% file: tables/table_MLP_CPU_Inference.tex
\begin{table}[htbp]
\centering
\caption{Duration, Emission, Energy and Performance in Inference phase for Multi Perceptron Model deep learning model with CPU}
\label{table:MLP_CPU_inference}
\resizebox{\linewidth}{!}{%
\begin{tblr}{
  column{3} = {c},
  column{10} = {c},
  cell{1}{1} = {r=2}{},
  cell{1}{2} = {r=2}{},
  cell{1}{3} = {r=2}{},
  cell{1}{4} = {c=2}{},
  cell{1}{6} = {c=2}{},
  cell{1}{8} = {c=2}{},
  cell{1}{10} = {r=2}{},
  cell{3}{4} = {c},
  cell{3}{5} = {c},
  cell{3}{6} = {c},
  cell{3}{7} = {c},
  cell{3}{8} = {c},
  cell{3}{9} = {c},
  cell{4}{4} = {c},
  cell{4}{5} = {c},
  cell{4}{6} = {c},
  cell{4}{7} = {c},
  cell{4}{8} = {c},
  cell{4}{9} = {c},
  cell{5}{4} = {c},
  cell{5}{5} = {c},
  cell{5}{6} = {c},
  cell{5}{7} = {c},
  cell{5}{8} = {c},
  cell{5}{9} = {c},
  cell{6}{4} = {c},
  cell{6}{5} = {c},
  cell{6}{6} = {c},
  cell{6}{7} = {c},
  cell{6}{8} = {c},
  cell{6}{9} = {c},
  cell{7}{4} = {c},
  cell{7}{5} = {c},
  cell{7}{6} = {c},
  cell{7}{7} = {c},
  cell{7}{8} = {c},
  cell{7}{9} = {c},
  cell{8}{4} = {c},
  cell{8}{5} = {c},
  cell{8}{6} = {c},
  cell{8}{7} = {c},
  cell{8}{8} = {c},
  cell{8}{9} = {c},
  cell{9}{4} = {c},
  cell{9}{5} = {c},
  cell{9}{6} = {c},
  cell{9}{7} = {c},
  cell{9}{8} = {c},
  cell{9}{9} = {c},
  cell{10}{4} = {c},
  cell{10}{5} = {c},
  cell{10}{6} = {c},
  cell{10}{7} = {c},
  cell{10}{8} = {c},
  cell{10}{9} = {c},
  cell{11}{4} = {c},
  cell{11}{5} = {c},
  cell{11}{6} = {c},
  cell{11}{7} = {c},
  cell{11}{8} = {c},
  cell{11}{9} = {c},
  cell{12}{4} = {c},
  cell{12}{5} = {c},
  cell{12}{6} = {c},
  cell{12}{7} = {c},
  cell{12}{8} = {c},
  cell{12}{9} = {c},
  vline{4,6,8,10} = {3-12}{},
  hline{1,13} = {-}{0.08em},
  hline{3} = {-}{},
}
Model   & Config          & Parameters & Time (Sec) &      & CO2 equivalent (Kg) &          & Energy Consumed (KW) &          & F1 Macro \\
        &                 &            & Mean       & Std  & Mean                & Std      & Mean                 & Std      &          \\
MLP\_1  & 100             & 2903       & 6.94       & 1.31 & 2.00E-05            & 4.00E-06 & 9.10E-05             & 1.70E-05 & 0.6578   \\
MLP\_2  & 100, 70         & 9883       & 6.59       & 1.04 & 1.90E-05            & 3.00E-06 & 8.60E-05             & 1.40E-05 & 0.5992   \\
MLP\_3  & 100, 70, 50     & 13373      & 5.80       & 0.86 & 1.70E-05            & 2.00E-06 & 7.60E-05             & 1.10E-05 & 0.5992   \\
MLP\_4  & 100, 70, 50, 20 & 14303      & 7.39       & 1.32 & 2.10E-05            & 4.00E-06 & 9.70E-05             & 1.70E-05 & 0.7031   \\
MLP\_5  & 50              & 1453       & 6.30       & 1.14 & 1.80E-05            & 3.00E-06 & 8.30E-05             & 1.50E-05 & 0.7144   \\
MLP\_6  & 50, 50          & 4003       & 7.10       & 1.59 & 2.00E-05            & 5.00E-06 & 9.30E-05             & 2.10E-05 & 0.5992   \\
MLP\_7  & 200             & 5803       & 6.83       & 1.18 & 2.00E-05            & 3.00E-06 & 9.00E-05             & 1.50E-05 & 0.7143   \\
MLP\_8  & 200, 100        & 25603      & 8.76       & 1.70 & 2.50E-05            & 5.00E-06 & 1.15E-04             & 2.20E-05 & 0.5978   \\
MLP\_9  & 50, 40, 30, 20  & 5253       & 6.57       & 1.03 & 1.90E-05            & 3.00E-06 & 8.60E-05             & 1.40E-05 & 0.6598   \\
MLP\_10 & 40, 10          & 1483       & 6.56       & 1.38 & 1.90E-05            & 4.00E-06 & 8.60E-05             & 1.80E-05 & 0.5992   
\end{tblr}
}
\end{table}

%% file: tables/table_MLP_GPU_Inference.tex
\definecolor{MineShaft}{rgb}{0.129,0.129,0.129}
\begin{table}[htbp]
\caption{Duration, Emission, Energy and Performance in Inference phase for Multi Perceptron Model deep learning model with GPU}
\label{table:MLP_GPU_inference}
\resizebox{\linewidth}{!}{%
\centering
\begin{tblr}{
  row{1} = {c},
  row{2} = {c},
  cell{1}{1} = {r=2}{},
  cell{1}{2} = {r=2}{},
  cell{1}{3} = {r=2}{},
  cell{1}{4} = {c=2}{},
  cell{1}{6} = {c=2}{},
  cell{1}{8} = {c=2}{},
  cell{1}{10} = {r=2}{},
  cell{3}{3} = {c,},
  cell{3}{10} = {c},
  cell{4}{3} = {c,},
  cell{4}{10} = {c},
  cell{5}{3} = {c,},
  cell{5}{10} = {c},
  cell{6}{3} = {c,},
  cell{6}{10} = {c},
  cell{7}{3} = {c,},
  cell{7}{10} = {c},
  cell{8}{3} = {c,},
  cell{8}{10} = {c},
  cell{9}{3} = {c,},
  cell{9}{10} = {c},
  cell{10}{3} = {c,},
  cell{10}{10} = {c},
  cell{11}{3} = {c,},
  cell{11}{10} = {c},
  cell{12}{3} = {c,},
  cell{12}{10} = {c},
  vline{4,6,8,10} = {3-12}{},
  hline{1,3,13} = {-}{},
}
\textbf{Model} & \textbf{Config} & \textbf{Parameters} & \textbf{Time (Sec)} &              & \textbf{CO2 equivalent (Kg)} &              & \textbf{Energy Consumed (KW)} &              & \textbf{F1 Macro} \\
               &                 &                     & \textbf{Mean}       & \textbf{Std} & \textbf{Mean}                & \textbf{Std} & \textbf{Mean}                 & \textbf{Std} &                   \\
MLP\_1         & 100             & 2903                & 4.46                & 0.75         & 2.70E-05                     & 4.49E-06     & 7.00E-05                      & 1.18E-05     & 0.6578            \\
MLP\_2         & 100, 70         & 9883                & 2.85                & 0.04         & 1.70E-05                     & 2.45E-07     & 4.50E-05                      & 6.45E-07     & 0.5992            \\
MLP\_3         & 100, 70, 50     & 13373               & 4.77                & 0.78         & 2.80E-05                     & 4.68E-06     & 7.50E-05                      & 1.23E-05     & 0.5992            \\
MLP\_4         & 100, 70, 50, 20 & 14303               & 5.17                & 0.44         & 3.10E-05                     & 2.67E-06     & 8.20E-05                      & 7.03E-06     & 0.7031            \\
MLP\_5         & 50              & 1453                & 5.07                & 0.88         & 3.00E-05                     & 5.33E-06     & 8.00E-05                      & 1.41E-05     & \textbf{0.7144}   \\
MLP\_6         & 50, 50          & 4003                & 4.12                & 0.58         & 2.50E-05                     & 3.38E-06     & 6.50E-05                      & 8.90E-06     & 0.5992            \\
MLP\_7         & 200             & 5803                & 2.84                & 0.06         & 1.70E-05                     & 4.03E-07     & 4.50E-05                      & 1.06E-06     & 0.7143            \\
MLP\_8         & 200, 100        & 25603               & 4.22                & 0.58         & 2.50E-05                     & 3.40E-06     & 6.60E-05                      & 8.96E-06     & 0.5978            \\
MLP\_9         & 50, 40, 30, 20  & 5253                & 5.71                & 0.91         & 3.40E-05                     & 5.47E-06     & 9.00E-05                      & 1.44E-05     & 0.6598            \\
MLP\_10        & 40, 10          & 1483                & 4.70                & 0.74         & 2.80E-05                     & 4.43E-06     & 7.40E-05                      & 1.17E-05     & 0.5992            
\end{tblr}
}
\end{table}

%% file: tables/table_quantization.tex
\begin{table}[htbp]
\caption{Comparison of MLP\_5 model in quantizated int8 and float 32 CPU and GPU precision at Inference phase.}
\resizebox{\linewidth}{!}{%
\centering
\begin{tblr}{
  column{2} = {c},
  cell{1}{1} = {r=2}{},
  cell{1}{2} = {r=2}{},
  cell{1}{3} = {c=2}{},
  cell{1}{5} = {c=2}{},
  cell{1}{7} = {c=2}{},
  cell{1}{9} = {r=2}{},
  cell{1}{10} = {r=2}{},
  cell{3}{3} = {c},
  cell{3}{4} = {c},
  cell{3}{5} = {c},
  cell{3}{6} = {c},
  cell{3}{7} = {c},
  cell{3}{8} = {c},
  cell{3}{9} = {c},
  cell{3}{10} = {c},
  cell{4}{3} = {c},
  cell{4}{4} = {c},
  cell{4}{5} = {c},
  cell{4}{6} = {c},
  cell{4}{7} = {c},
  cell{4}{8} = {c},
  cell{4}{9} = {c},
  cell{4}{10} = {c},
  cell{5}{3} = {c},
  cell{5}{4} = {c},
  cell{5}{5} = {c},
  cell{5}{6} = {c},
  cell{5}{7} = {c},
  cell{5}{8} = {c},
  cell{5}{9} = {c},
  cell{5}{10} = {c},
  vline{2-3,5,7,9-10} = {3-5}{},
  hline{1,6} = {-}{0.08em},
  hline{3} = {-}{},
}
\textbf{Model} & \textbf{Size (KB)} & \textbf{Time (Sec)} &              & \textbf{CO2 equivalent (Kg)} &              & \textbf{Energy Consumed (KW)} &              & \textbf{F1 Macro} & \textbf{Val Loss} \\
               &                    & \textbf{Mean}       & \textbf{Std} & \textbf{Mean}                & \textbf{Std} & \textbf{Mean}                 & \textbf{Std} &                   &                   \\
int8\_cpu      & 4.6                & 5.63                & 0.88         & 5.66E-05                     & 9.00E-06     & 1.16E-04                      & 1.80E-05     & 0.7223            & 0.00264764        \\
fp32\_cpu      & 7.6                & 6.30                & 1.14         & 6.32E-05                     & 5.33E-06     & 1.29E-04                      & 1.41E-05     & 0.7223            & 0.00264765        \\
fp32\_gpu      & 7.6                & 8.34                & 0.42         & 8.40E-05                     & 1.89E-06     & 1.72E-04                      & 9.00E-06     & 0.7223            & 0.00264765        
\end{tblr}
}
\end{table}